\documentclass[11pt]{article}

\usepackage[preprint]{acl}

\usepackage{times}
\usepackage{latexsym}
\usepackage[T1]{fontenc}
\usepackage[utf8]{inputenc}
\usepackage{microtype}
\usepackage{inconsolata}
\usepackage{url}
\usepackage{graphicx}
\usepackage{multirow}
\usepackage{amsmath}
\usepackage{amssymb}
\usepackage{booktabs}
\usepackage{xspace}
\usepackage{tikz}
\usepackage{algorithm}
\usepackage{algpseudocode}
\usetikzlibrary{positioning,arrows.meta,decorations.pathreplacing,calc,fit,shapes.geometric,backgrounds}

\newcommand{\name}{BayLing-Duplex\xspace}

\title{\name: Native Full-Duplex Speech Dialogue \\ with a Single Autoregressive LLM}

\author{
    Qingkai Fang$^{1,2,3}$,
    Shoutao Guo$^{1,2,3}$,
    Yang Feng$^{1,2,3}$\thanks{Corresponding author: Yang Feng.} \\
    \textsuperscript{\rm1}Key Laboratory of Intelligent Information Processing \\ Institute of Computing Technology, Chinese Academy of Sciences (ICT/CAS) \\
    \textsuperscript{\rm2}Key Laboratory of AI Safety, Chinese Academy of Sciences \\
    \textsuperscript{\rm3}University of Chinese Academy of Sciences, Beijing, China \\
    {$\;\:$\texttt{\{fangqingkai21b,guoshoutao22z,fengyang\}@ict.ac.cn}}
}

\begin{document}
\maketitle

\begin{abstract}
Real-time, full-duplex speech interaction is a key feature of next-generation spoken chatbots, allowing the model to listen and speak at the same time and to handle natural phenomena such as overlap, hesitation, and barge-in. Existing speech language models (SpeechLMs) such as LLaMA-Omni~\citep{fang2024llamaomni} and GLM-4-Voice~\citep{glm4voice} are still turn-based and rely on an external Voice Activity Detection (VAD) module to mark the end of the user's turn, which fundamentally limits their interactive ability. In this paper, we introduce \name, a native full-duplex SpeechLM where a single autoregressive LLM decides when to listen, when to speak, and when to stop, with no auxiliary turn-taking module. The design adds only a few special tokens to the standard vocabulary, so it transfers across LLMs and reuses existing training and serving stacks with no architectural adaptation. Starting from the public GLM-4-Voice checkpoint and using only 400K full-duplex samples for fine-tuning followed by a lightweight DPO stage, \name reaches 92\% turn-taking success and 100\% interruption success on InstructS2S-Eval, while improving the speech-response score from 2.17 to 3.39 over Moshi~\citep{defossez2024moshi}. \name also matches or surpasses its turn-based counterpart on Llama Questions, Web Questions, and Alpaca-Eval, showing that simultaneous listen-and-speak modeling does not sacrifice response quality.\footnote{Code and models are available at \url{https://github.com/BayLing-Models/BayLing-Duplex}.}
\end{abstract}

\section{Introduction}
\label{sec:intro}

Speech, as a critical interface for human-computer interaction, can enhance user experience compared with text. In recent years, with the rapid development of large language models (LLMs), building intelligent spoken chatbots has attracted widespread attention from both academia and industry. GPT-4o~\citep{gpt4o} enables real-time, intelligent, and natural speech interaction, marking a step toward more natural human-computer interaction.

The traditional approach is a cascaded pipeline of automatic speech recognition (ASR), an LLM, and text-to-speech (TTS) synthesis. While straightforward, the cascaded design accumulates errors across stages, suffers from high response latency, and discards paralinguistic information in the input speech. To address these limitations, end-to-end SpeechLMs have gained attention, using a single unified model to process speech input and output. They can be categorized into \textit{native} SpeechLMs that discretize speech into tokens and extend the LLM vocabulary~\citep{zhang2023speechgpt, glm4voice, defossez2024moshi}, and \textit{modular} SpeechLMs that incorporate a speech encoder and a speech decoder around the LLM~\citep{fang2024llamaomni, fang2025llamaomni2, wang2024freezeomni}. Despite different architectures, both families predominantly assume a \textit{turn-based} interaction: the model consumes one segmented user utterance and emits a single response.

Deployment therefore requires a front-end Voice Activity Detection (VAD) module to mark the end of the user's turn. The turn-based assumption has two intrinsic limitations. First, the system behavior is bounded by the VAD's accuracy: false positives cut the user off mid-sentence and false negatives delay the response, since acoustic VAD has no access to dialogue semantics. Second, the turn-based abstraction discards interaction patterns that pervade real conversation, including mid-utterance pauses that should not be mistaken for end-of-turn, user barge-in that should preempt the current response, and short backchannels that should not trigger a full reply. Outsourcing these decisions to a small front-end module places a hard ceiling on the system's interactive ability. \textit{Full-duplex} SpeechLMs address these issues by listening and speaking continuously, deciding internally when to talk~\citep{nguyen2022dgslm, defossez2024moshi, zhang2024omniflatten}. However, native full-duplex training typically requires millions of hours of pretraining and tens of thousands of hours of paired full-duplex dialogue data~\citep{defossez2024moshi}, which is beyond the reach of most academic teams. In this paper, we explore an alternative: converting a strong turn-based SpeechLM into a competitive full-duplex one with a small, structured fine-tuning recipe. The conversion is non-trivial, since the model must consume the user's incoming speech while emitting its own response, and make every turn-taking decision at the same time scale as speech tokens.

In this paper, we propose \name, a native full-duplex SpeechLM in which a single autoregressive LLM jointly handles user-speech understanding, dialogue-state decisions, and assistant-speech generation through a multi-channel interleaved sequence (Figure~\ref{fig:duplex}). \name takes GLM-4-Voice~\citep{glm4voice} as its backbone, integrating a speech tokenizer, an LLM, and a speech decoder; we introduce no new modules or auxiliary heads on top of the GLM-4-Voice backbone; the only addition is four special dialogue-state tokens that share the standard token vocabulary. As a result, the design transfers to any autoregressive LLM and runs on off-the-shelf LLM training and serving frameworks without any architectural adaptation. Three streams -- user speech, assistant text, and assistant speech -- are tokenized at the same frame rate and interleaved block by block, and four dialogue-state tokens in the text channel encode silence, reply onset, text completion, and speech completion. With this layout, every turn-taking and interruption decision reduces to ordinary next-token prediction over GLM-4-Voice's standard vocabulary. We start from the publicly released GLM-4-Voice checkpoint and fine-tune it on 400K full-duplex samples, followed by a lightweight Direct Preference Optimization (DPO)~\citep{rafailov2023dpo} stage targeting turn-taking and barge-in timing. Experimental results show that \name reaches 92\% turn-taking success rate and 100\% interruption success rate on InstructS2S-Eval, while improving the speech-response score from 2.17 to 3.39 over Moshi~\citep{defossez2024moshi}. On full-duplex spoken question answering, \name reaches 46.0\%/18.1\% accuracy on Llama Questions and Web Questions, significantly outperforming Moshi's 21.0\%/9.2\%, and the duplex model is on par with or stronger than its turn-based counterpart on three standard spoken benchmarks.

\section{\name}
\label{sec:method}

\begin{figure*}[t]
\centering
\includegraphics[width=\textwidth]{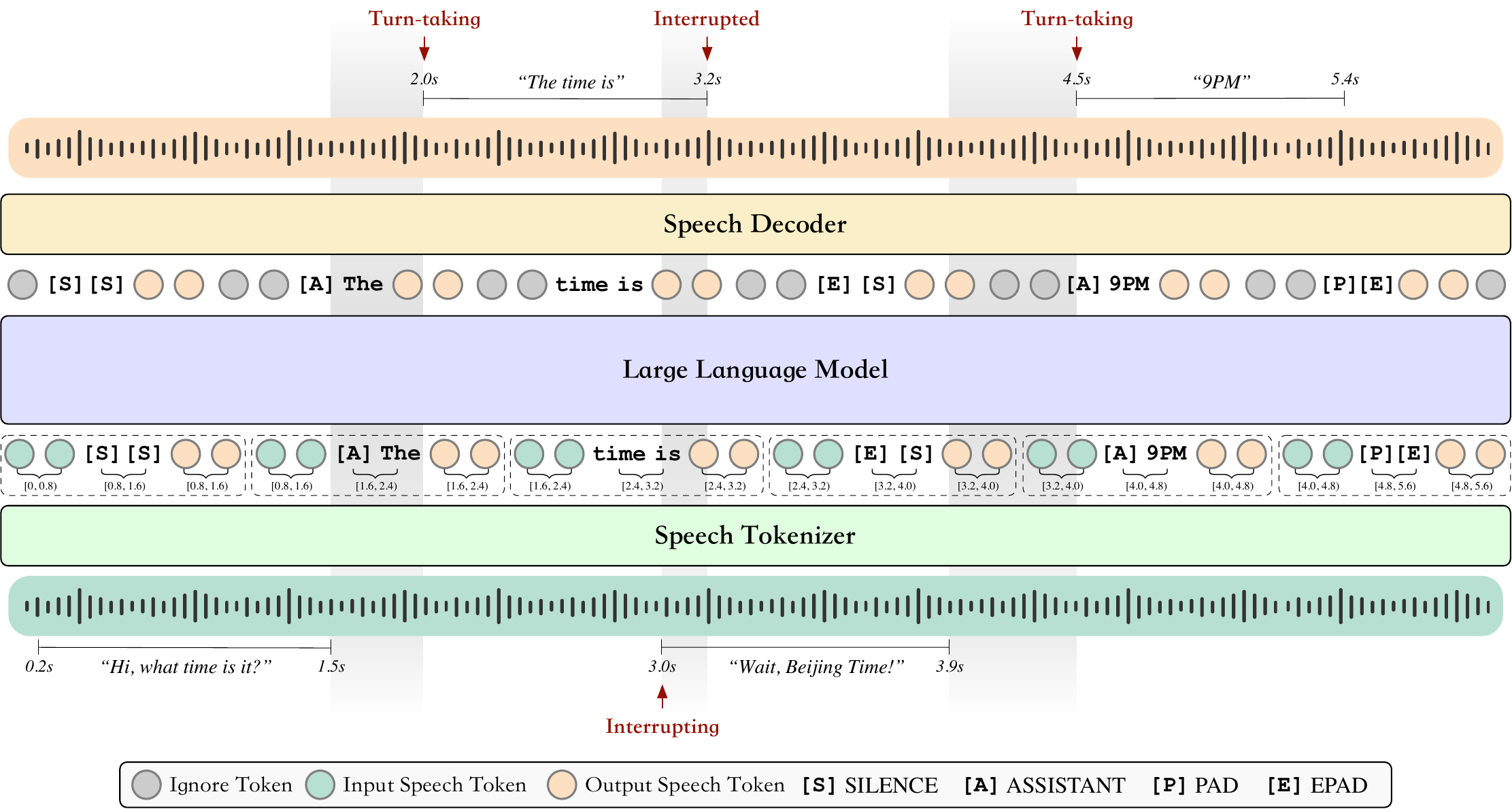}
\caption{Multi-channel interleaved sequence in BayLing-Duplex. The user speech, assistant text and assistant speech channels are interleaved block-by-block at a fixed $N{:}M{:}N$ ratio; here $N{=}M{=}2$ for clarity (we use $N{=}10$, $M{=}5$ in practice). The text channel embeds the dialogue-state tokens \texttt{[SILENCE]} (\texttt{[S]}), \texttt{[ASSISTANT]} (\texttt{[A]}), \texttt{[PAD]} (\texttt{[P]}) and \texttt{[EPAD]} (\texttt{[E]}). The illustrated dialogue starts with the user asking ``Hi, what time is it?''; the assistant takes the turn at $2.0$\,s with ``The time is\ldots'', is interrupted at $3.0$\,s by ``Wait, Beijing Time!'', and re-starts with ``9PM'' at $4.5$\,s. Turn-taking, being interrupted, and re-starting are all expressed as ordinary next-token prediction over the standard LLM vocabulary.}
\label{fig:duplex}
\end{figure*}

In this section, we introduce the model architecture of \name. As shown in Figure~\ref{fig:duplex}, we use GLM-4-Voice~\citep{glm4voice} as the backbone, which integrates a speech tokenizer, an LLM, and a speech decoder. The speech tokenizer is a modified Whisper-large-v3~\citep{whisper} encoder with a vector quantizer that turns 16\,kHz waveforms into discrete tokens at $f_s = 12.5$\,Hz (one token per 80\,ms); the LLM is a 9B-parameter decoder-only Transformer initialized from GLM-4-9B~\citep{glm2024chatglm} with the speech tokens added to its vocabulary; and the speech decoder is a flow-matching~\citep{lipman2023flow} model followed by a HiFi-GAN~\citep{hifigan} vocoder, both adapted from CosyVoice~\citep{du2024cosyvoice}. The core innovation of \name is the multi-channel interleaved sequence layout, which realizes full-duplex behavior without introducing any new modules or auxiliary heads.

\subsection{Multi-Channel Interleaved Sequence}
\label{sec:seq}

A full-duplex dialogue consists of a sequence of utterances by the user and the assistant, possibly with overlap to support barge-in. We organize this dialogue into a single multi-channel interleaved sequence as follows.

\paragraph{Two-Channel Audio Tokenization}
We synthesize two single-channel audio tracks of equal length: the user track is filled with user utterances (silence elsewhere) and the assistant track with assistant utterances. Both are tokenized by the speech tokenizer, yielding aligned sequences $\mathbf{X} = (x_1, \dots, x_{T_s})$ and $\mathbf{Y} = (y_1, \dots, y_{T_s})$. Silence is tokenized by the same encoder rather than replaced by a special token, preserving acoustic continuity. For each assistant utterance $k$, $\mathbf{w}_k$ denotes its textual content, and $s_k, e_k$ denote its start and end times (in seconds), respectively.

\paragraph{Block Structure}
The sequence is organized in $B$ blocks, each containing $N$ user-speech tokens, $M$ text tokens, and $N$ assistant-speech tokens:
\begin{equation}
\small
\text{Block }b\!:\;
\underbrace{x_{bN+1{:}(b+1)N}}_{\text{user speech}}
\,
\underbrace{\mathbf{z}_{bM+1{:}(b+1)M}}_{\text{text}}
\,
\underbrace{y_{bN+1{:}(b+1)N}}_{\text{assistant speech}}.
\end{equation}
The text channel $\mathbf{Z}=(z_1,\dots,z_{T_z})$ has length $T_z = T_s \cdot M/N$. The model is trained to predict the text and assistant-speech tokens autoregressively given the past sequence.

\paragraph{Block Size}
The block size $N$ controls a fundamental trade-off. With a small $N$, each block has too few text slots to express even a short sub-word, which produces jittery turn-taking and unstable response timing; with a large $N$, the minimum response latency exceeds the human-acceptability threshold, since the model can only respond at the granularity of one block~\citep{defossez2024moshi}. We choose $N=10$ and $M=5$ throughout the paper, giving $\Delta t = 0.8$\,s and 6.25 text tokens per second on average, close to the natural English speech rate of GLM-4-Voice during turn-based decoding. $N{=}10$ matches the typical English minimum-perceptible-latency threshold while keeping $\Delta t$ small enough for fluid turn-taking; we leave a systematic sweep over $N$ to future work.

\paragraph{Causal Shift}
At block $b$, the model has observed user tokens up to time $(b{+}1)\Delta t$, so the earliest assistant audio it can emit corresponds to that same instant. We therefore shift the assistant text and speech channels one block ahead of the user channel during training: text and assistant-speech tokens at block $b$ correspond to wall-clock window $[(b{+}1)\Delta t, (b{+}2)\Delta t)$. At inference, the output is played back with the same offset $\Delta t$ added.

\paragraph{Text-Channel Construction}
The text channel $\mathbf{Z}$ acts as an inner monologue: it never reaches the user, but conditions the assistant-speech tokens within the same block. $\mathbf{Z}$ is initialized with \texttt{[SILENCE]} everywhere and overwritten by each assistant utterance $k$. Its boundary indices in the text channel are
\begin{align}
\small
j_k^{\text{ast}} &= \lfloor (s_k - \Delta t) f_s \rfloor \cdot \tfrac{M}{N} - 1, \\
\small
j_k^{\text{epad}} &= \lceil (e_k - \Delta t) f_s \rceil \cdot \tfrac{M}{N},
\end{align}
with the textual content $\mathbf{w}_k$ filling positions from $j_k^{\text{ast}}{+}1$. The text channel embeds four dialogue-state tokens that encode the high-level state of the dialogue:
\begin{itemize}
\setlength\itemsep{0pt}
\item \texttt{[SILENCE]}: the assistant should stay silent;
\item \texttt{[ASSISTANT]}: the start of an assistant reply;
\item \texttt{[PAD]}: the textual content has been written but the corresponding speech is still being emitted;
\item \texttt{[EPAD]}: both the text and the speech of the current reply are complete.
\end{itemize}
When the text channel emits \texttt{[SILENCE]} the assistant-speech tokens correspond to silence; when it emits \texttt{[ASSISTANT]} followed by content, the assistant-speech tokens encode the corresponding utterance. With this layout, all dialogue-state decisions reduce to next-token prediction over GLM-4-Voice's standard vocabulary, requiring no extra classification head, attention-mask trick, or state machine.

\subsection{Training}
\label{sec:train}

We start from GLM-4-Voice's publicly released checkpoint, which has already been pretrained on millions of hours of speech-text data and supervised fine-tuned on turn-based dialogue. Two further stages are applied.

\paragraph{Stage I: Supervised Fine-Tuning}
The user-speech channel $\mathbf{X}$ is conditioning only and contributes no loss; the cross-entropy is evaluated only at text-channel and assistant-speech positions, with the supervised set
\begin{equation}
\small
\mathcal{V} = \{i : s_i \in \mathbf{Z} \cup \mathbf{Y}\}, \quad
\ell_i = -\log \pi_\theta(s_i \mid \mathbf{s}_{<i}).
\end{equation}
\texttt{[SILENCE]} dominates a typical sequence while \texttt{[ASSISTANT]} appears only once per turn, so we aggregate the per-position losses with per-token weights $\omega_i$ to keep the rare role tokens from being drowned out:
\begin{equation}
\small
\mathcal{L}_{\text{SFT}} = \frac{\sum_{i \in \mathcal{V}} \omega_i \ell_i}{\sum_{i \in \mathcal{V}} \omega_i}.
\end{equation}
We tune two key weights: $\omega_{\text{sil}}$ for \texttt{[SILENCE]} and $\omega_{\text{role}}$ for \texttt{[ASSISTANT]}/\texttt{[EPAD]}; we write $\mathcal{L}_{\text{SFT}}(\mathbf{s})$ when this loss is evaluated on a specific sequence $\mathbf{s}$. Ablations are reported in Section~\ref{sec:ablation}.

\paragraph{Stage II: Direct Preference Optimization}
Stage I teaches the layout but only weakly optimizes temporal decisions. We construct preference pairs whose positive examples are the SFT data and whose negatives differ \emph{only} in timing; the construction is detailed in Section~\ref{sec:data}. The training objective combines DPO with an auxiliary SFT term that prevents catastrophic forgetting of generation quality:
\begin{equation}
\small
\mathcal{L} = \mathcal{L}_{\text{DPO}} + \lambda_{\text{ftx}} \cdot \mathcal{L}_{\text{SFT}}(\mathbf{s}^+),
\end{equation}
\begin{equation}
\small
\mathcal{L}_{\text{DPO}} \!=\! -\log\!\sigma\!\!\left(\!\beta\!\left[\log\tfrac{\pi_\theta(\mathbf{s}^+)}{\pi_{\text{ref}}(\mathbf{s}^+)} \!-\! \log\tfrac{\pi_\theta(\mathbf{s}^-)}{\pi_{\text{ref}}(\mathbf{s}^-)}\right]\!\right),
\end{equation}
where $\pi_{\text{ref}}$ is the Stage I checkpoint.

\subsection{Inference}
\label{sec:inference}

\begin{algorithm}[t]
\small
\caption{Inference of \name.}
\label{algo:fullduplex_infer}
\begin{algorithmic}[1]
\Require live user-speech stream; block sizes $N$, $M$; causal offset $\Delta t$
\Ensure assistant-speech waveform
\State $b \gets 0$, history $\mathbf{S} \gets ()$
\While{dialogue is active}
    \State receive $N$ user-speech tokens $\mathbf{x}_b$ from the stream
    \State $\mathbf{S} \gets \mathbf{S} \oplus \mathbf{x}_b$
    \For{$j = 1, \ldots, M$} \Comment{text channel}
        \State $z_j \sim \pi_\theta(\cdot \mid \mathbf{S})$, mask to text + state tokens
        \State $\mathbf{S} \gets \mathbf{S} \oplus z_j$
    \EndFor
    \For{$j = 1, \ldots, N$} \Comment{assistant speech}
        \State $y_j \sim \pi_\theta(\cdot \mid \mathbf{S})$, mask to speech tokens
        \State $\mathbf{S} \gets \mathbf{S} \oplus y_j$
    \EndFor
    \State decode $\{y_1,\dots,y_N\}$ and play at $(b{+}1)\Delta t$
    \State $b \gets b + 1$
\EndWhile
\end{algorithmic}
\end{algorithm}

During inference, decoding proceeds block by block, as summarized in Algorithm~\ref{algo:fullduplex_infer}. The text-channel slots are masked to text-and-state tokens and the assistant-speech slots to speech tokens; without masking, the LLM occasionally emits cross-channel tokens that corrupt the speech decoder's input. During training no mask is applied because the cross-entropy loss naturally suppresses incorrect token types. During silence the user channel still receives the live waveform, which the tokenizer maps to its silence token; user input is never zero-padded artificially. When the user barges in mid-block, the in-flight assistant-speech tokens finish generating before the next block re-conditions on the new user audio, keeping decoding strictly autoregressive.

\section{Data Construction}
\label{sec:data}

In this section, we describe how we construct the full-duplex training data. We build upon the multi-turn speech-to-speech dialogue corpus introduced in \citet{fang2025llamaomni2}, which contains 200K samples derived from the Alpaca and UltraChat datasets through rewriting with Llama-3.3-70B-Instruct and synthesized into speech with CosyVoice's zero-shot voice cloning~\citep{du2024cosyvoice}. The user instructions are synthesized with diverse voices via voice cloning, while the assistant responses use a uniform voice; this preserves voice diversity across dialogues and consistency within a dialogue.

Each multi-turn dialogue is then converted into the multi-channel interleaved format for two full-duplex scenarios. For \textbf{turn-taking}, a 0.8\,s gap is inserted from the end of the user's utterance to the start of the assistant's response, and the gap from the end of the assistant's response to the start of the next user utterance is drawn from $\mathrm{Uniform}(0.5, 3.0)$\,s. For \textbf{interruption}, the user re-enters at a random point during the assistant's response, and the assistant stops after a small reaction delay $\delta_{\text{react}} \sim \mathrm{Uniform}(0.8, 2.0)$\,s. We generate 200K full-duplex samples for each scenario and mix them in a 1:1 ratio during training.

For DPO, we reuse the same SFT samples as positive examples and synthesize negatives by perturbing only the timing of the assistant. For \textbf{turn-taking}, the negative replaces the 0.8\,s gap with a value drawn from $\mathrm{Uniform}(2,5)$\,s, which forces the SFT model to over-predict \texttt{[SILENCE]} after the user finishes speaking. For \textbf{interruption}, the negative replaces the reaction delay with a value drawn from $\mathrm{Uniform}(3, 5)$\,s, so that the assistant continues to speak well after the user has barged in. Each positive is paired with one negative; positive and negative share the same user-channel audio and textual content $\mathbf{w}_k$, so the DPO objective is forced to focus its update on the dialogue-state tokens and not on textual content, which is essential for preserving response quality during the preference-optimization stage.

\section{Experiments}
\label{sec:exp}

\subsection{Experimental Setup}
\label{sec:setup}

\paragraph{Model Configuration}
We use the GLM-4-Voice checkpoint as the backbone, with $N{=}10$ and $M{=}5$ (block duration $\Delta t = 0.8$\,s). The LLM is fully fine-tuned, while the speech tokenizer and speech decoder are frozen. We add no new parameters or auxiliary heads.

\paragraph{Training Details}
Stage I (SFT) is trained on the 400K full-duplex dialogues described in Section~\ref{sec:data} for one epoch with batch size 32 and a peak learning rate of $1\!\times\!10^{-5}$, using a cosine schedule with 10\% warm-up. Stage II (DPO) runs for 200 steps with a peak learning rate of $3\!\times\!10^{-7}$, $\beta\!=\!0.5$, and $\lambda_{\text{ftx}}\!=\!0.5$, using a cosine schedule with 5\% warm-up. Both stages are trained with the LLaMA-Factory codebase~\citep{zheng2024llamafactory}.

\subsection{Evaluation}
\label{sec:eval}

We evaluate \name on three tasks: spoken question answering, full-duplex turn-taking, and full-duplex interruption. In all experiments we sample from the LLM with temperature 0.8. The synthesized assistant audio is transcribed by Whisper-large-v3~\citep{whisper} and segmented by Silero VAD~\citep{silerovad}.

\paragraph{Spoken Question Answering}
The spoken question answering task feeds a spoken question directly to the full-duplex model with no external VAD, and checks whether the reference answer appears in the model's response. We evaluate on Llama Questions~\citep{nachmani2024spoken} (300 items) and Web Questions~\citep{berant2013semantic} (2032 items, synthesized into speech by CosyVoice).

\paragraph{Turn-Taking}
For turn-taking, we follow LLaMA-Omni~\citep{fang2024llamaomni} and use \textit{InstructS2S-Eval}, 199 spoken instructions filtered from the \textit{helpful\_base} and \textit{vicuna} subsets of Alpaca-Eval~\citep{alpacaeval}. We feed each instruction to the duplex model in real time and measure when the assistant starts replying after the user finishes speaking, as well as the quality of the reply.

\paragraph{Interruption}
For interruption, we pair adjacent items from InstructS2S-Eval into 199 two-utterance audios where the second utterance starts during the first response. We measure how quickly the model stops the current response when interrupted, and how relevant the new reply is to the second question.

All timing metrics are computed on the synthesized assistant audio: we run Silero VAD~\citep{silerovad} on the waveform to obtain its non-silence segments, define $t_\text{user-end}$ as the right edge of the last non-silence frame in the synthesized user audio, $t_\text{assistant-start}$ as the start of the assistant's first non-silence segment, and $t_\text{stop}$ as the right edge of the assistant's last non-silence segment that follows a barge-in (i.e., when the assistant falls silent after being interrupted). The evaluation pipeline never inspects the model's text channel or special tokens, and Silero VAD is used only for evaluation, not for inference.

We use the following metrics.

\textbf{TT SR@3s}: turn-taking success rate, defined as the fraction of test items for which the assistant starts replying within 3\,s of the user's end.

\textbf{S2S Score}: a 1--5 GPT-4o~\citep{gpt4o} judgment on the transcribed assistant reply, considering helpfulness, relevance, fluency, and suitability for speech interaction.

\textbf{Overlap (Ovl)}: the gap, in seconds, from the user's barge-in to the assistant's stop; lower is better.

\textbf{ISR@2s}: the interruption success rate, defined as the fraction of test items whose overlap is at most 2\,s.

\textbf{Q2 S2S}: the S2S Score on the assistant's reply to the second (interrupting) question, used to measure whether the model produces a relevant new response after being interrupted.

For spoken QA, we report exact-match accuracy. Accuracy is computed by case-insensitive substring match between the reference answer and the Whisper transcription of the assistant's audio.

\paragraph{Baseline}
We compare \name with Moshi~\citep{defossez2024moshi}, a state-of-the-art native full-duplex SpeechLM with parallel audio streams and an Inner Monologue text channel. We use the publicly released Moshika checkpoint.

\subsection{Main Results}
\label{sec:main}

\begin{table}[t]
    \centering
    \setlength{\tabcolsep}{6pt}
    \resizebox{\columnwidth}{!}{%
    \begin{tabular}{lcc}
        \toprule
        \multirow{2}{*}{\textbf{Model}} & \textbf{Llama Q.}$\uparrow$ & \textbf{Web Q.}$\uparrow$ \\
        & ($N{=}300$) & ($N{=}2032$) \\
        \midrule
        Moshi & 21.0 & 9.2 \\
        BayLing-Duplex (SFT) & 44.3 & 18.0 \\
        BayLing-Duplex (+DPO) & \textbf{46.0} & \textbf{18.1} \\
        \bottomrule
    \end{tabular}}
    \caption{Full-duplex spoken QA accuracy (\%). The audio is fed directly to the duplex model with no external VAD.}
    \label{tab:fullduplex_qa}
\end{table}

\begin{table*}[t]
    \centering
    \setlength{\tabcolsep}{8pt}
    \begin{tabular}{lccccc}
        \toprule
        \multirow{2}{*}{\textbf{Model}} & \multicolumn{2}{c}{\textbf{Turn-taking}} & \multicolumn{3}{c}{\textbf{Interruption}} \\
        \cmidrule(lr){2-3} \cmidrule(lr){4-6}
        & \textbf{SR@3s$\uparrow$} & \textbf{S2S$\uparrow$} & \textbf{Overlap (s)$\downarrow$} & \textbf{ISR@2s$\uparrow$} & \textbf{Q2 S2S$\uparrow$} \\
        \midrule
        Moshi & 71.9 & 2.17 & 2.07 & 81.9 & 2.45 \\
        \name\ (SFT) & 88.9 & 3.23 & 1.51 & 91.4 & 2.95 \\
        \name\ (+DPO) & \textbf{92.0} & \textbf{3.39} & \textbf{1.10} & \textbf{100.0} & \textbf{3.27} \\
        \bottomrule
    \end{tabular}
    \caption{Main results on full-duplex turn-taking and interruption on InstructS2S-Eval (199 spoken instructions). SR@3s: turn-taking success rate at 3\,s; S2S: GPT-4o speech-response score; Overlap: gap from user barge-in to assistant stop; ISR@2s: interruption success rate at 2\,s; Q2 S2S: speech-response score on the assistant's reply to the second (interrupting) question.}
    \label{tab:fullduplex_main}
\end{table*}

\paragraph{Spoken Question Answering}
Table~\ref{tab:fullduplex_qa} reports spoken-QA accuracy in the full-duplex setting, where the spoken question is fed directly to the duplex model and the model itself decides when to reply. We observe that: (1) \name (SFT) reaches 44.3\%/18.0\% on Llama/Web Questions, significantly outperforming Moshi's 21.0\%/9.2\% even before DPO. (2) DPO further improves accuracy to 46.0\%/18.1\%, indicating that better timing also yields better content. (3) The improvement is consistent across both benchmarks, suggesting that the multi-channel layout preserves the content-modeling capability of the GLM-4-Voice backbone.

\paragraph{Turn-Taking and Interruption}
Table~\ref{tab:fullduplex_main} shows turn-taking and interruption results. We observe that: (1) The SFT model already reaches 88.9\% TT SR@3s and a 3.23 S2S Score, significantly outperforming Moshi (71.9\%, 2.17). (2) DPO pushes TT SR@3s to 92.0\% and the S2S Score to 3.39, exceeding Moshi's 2.17 by 1.22 points. (3) The interruption gain is even larger: Overlap drops from 2.07\,s (Moshi) to 1.51\,s (SFT) and 1.10\,s (+DPO); ISR@2s climbs from 81.9\% to 100\%; and Q2 S2S rises from 2.45 to 3.27. Interruption benefits the most because the negatives in DPO directly postpone the \texttt{[EPAD]} token.

\subsection{Ablation Study}
\label{sec:ablation}

We conduct ablation studies to understand the contribution of each component.

\begin{table*}[t]
    \centering
    \setlength{\tabcolsep}{6pt}
    \begin{tabular}{ccccccc}
        \toprule
        \multirow{2}{*}{$\omega_{\text{role}}$} & \multirow{2}{*}{$\omega_{\text{sil}}$} & \multicolumn{2}{c}{\textbf{Turn-taking}} & \multicolumn{3}{c}{\textbf{Interruption}} \\
        \cmidrule(lr){3-4} \cmidrule(lr){5-7}
        & & \textbf{SR@3s$\uparrow$} & \textbf{S2S$\uparrow$} & \textbf{Ovl$\downarrow$} & \textbf{ISR@2s$\uparrow$} & \textbf{Q2 S2S$\uparrow$} \\
        \midrule
        1 & 1   & 60.3 & 3.19 & 1.79 & \textbf{100.0} & 2.82 \\
        1 & 0.1 & 82.4 & 3.13 & 1.53 & 89.6 & 2.78 \\
        10 & 1   & 73.9 & 3.02 & 1.53 & 88.5 & 2.81 \\
        10 & 0.1 & \textbf{88.9} & \textbf{3.23} & \textbf{1.51} & 91.4 & \textbf{2.95} \\
        \bottomrule
    \end{tabular}
    \caption{Token-weight ablation in Stage I. $\omega_{\text{role}}$ weights \texttt{[ASSISTANT]}/\texttt{[EPAD]}, $\omega_{\text{sil}}$ weights \texttt{[SILENCE]}.}
    \label{tab:fullduplex_weight}
\end{table*}

\begin{table*}[t]
    \centering
    \setlength{\tabcolsep}{6pt}
    \begin{tabular}{cccccc}
        \toprule
        \multirow{2}{*}{$\beta$} & \multirow{2}{*}{$\lambda_{\text{ftx}}$} & \multicolumn{2}{c}{\textbf{Turn-taking}} & \multicolumn{2}{c}{\textbf{Interruption}} \\
        \cmidrule(lr){3-4} \cmidrule(lr){5-6}
        & & \textbf{SR@3s$\uparrow$} & \textbf{S2S$\uparrow$} & \textbf{Ovl$\downarrow$} & \textbf{Q2 S2S$\uparrow$} \\
        \midrule
        0.1 & 0.3 & 86.4 & 3.36 & \textbf{1.01} & 3.09 \\
        0.1 & 0.5 & 89.4 & 3.31 & 1.05 & 3.07 \\
        0.1 & 1.0 & 89.4 & 3.30 & 1.10 & 3.10 \\
        0.3 & 0.3 & 89.4 & 3.38 & \textbf{1.01} & 3.06 \\
        0.3 & 0.5 & 87.9 & 3.30 & 1.05 & 3.13 \\
        0.3 & 1.0 & 87.4 & 3.36 & 1.11 & 3.22 \\
        0.5 & 0.3 & 90.1 & 3.35 & \textbf{1.01} & 3.17 \\
        0.5 & 0.5 & \textbf{92.0} & \textbf{3.39} & 1.10 & \textbf{3.27} \\
        0.5 & 1.0 & 89.4 & 3.37 & 1.12 & 3.11 \\
        \bottomrule
    \end{tabular}
    \caption{DPO hyperparameter ablation. $\beta$ is the KL coefficient and $\lambda_{\text{ftx}}$ is the auxiliary-SFT weight. ISR@2s = 100.0\% across all settings and is omitted.}
    \label{tab:fullduplex_dpo}
\end{table*}

\begin{table}[t]
    \centering
    \setlength{\tabcolsep}{4pt}
    \resizebox{\columnwidth}{!}{%
    \begin{tabular}{lccc}
        \toprule
        \multirow{2}{*}{\textbf{Model}} & \textbf{Llama Q.} & \textbf{Web Q.} & \textbf{Alpaca} \\
        & Acc.\ & Acc.\ & S2S \\
        \midrule
        Turn-based SFT & \textbf{45.3} & 15.9 & 3.16 \\
        BayLing-Duplex & 44.3 & \textbf{18.0} & \textbf{3.23} \\
        \bottomrule
    \end{tabular}}
    \caption{Response quality of \name vs.\ a turn-based SFT baseline trained on the same data and the same backbone.}
    \label{tab:fullduplex_turnbased}
\end{table}

\paragraph{Token Weights}
Table~\ref{tab:fullduplex_weight} shows the ablation on the per-token weights of the SFT loss. We observe that: (1) Uniform weighting ($\omega = 1$) collapses the model to near-permanent silence, with TT SR@3s of only 60.3\%. The 100\% ISR@2s in this row is a degenerate consequence: a model that almost never speaks needs no time to stop. (2) Reducing $\omega_{\text{sil}}$ to 0.1 alone raises TT SR to 82.4\%. (3) Raising $\omega_{\text{role}}$ to 10 with $\omega_{\text{sil}} = 0.1$ further pushes TT SR to 88.9\% and the speech score to 3.23. Both adjustments are needed: down-weighting \texttt{[SILENCE]} alone or up-weighting role tokens alone is insufficient, because the gradient is otherwise dominated by silence positions.

\paragraph{DPO Hyperparameters}
Table~\ref{tab:fullduplex_dpo} sweeps the Kullback--Leibler (KL) coefficient $\beta$ and the auxiliary-SFT coefficient $\lambda_{\text{ftx}}$. We observe that: (1) ISR@2s reaches 100\% across all settings, indicating that DPO is robust on interruption. (2) TT SR@3s and the S2S Score both peak at 92.0\%/3.39 with $\beta\!=\!0.5, \lambda_{\text{ftx}}\!=\!0.5$, which we use as the default. (3) $\lambda_{\text{ftx}}\!=\!1.0$ slightly degrades the DPO effect, while $\lambda_{\text{ftx}}\!=\!0.3$ recovers similar interaction quality but yields a lower S2S Score. (4) Lowering $\beta$ to 0.1 makes the model drift further from the SFT policy and produces a slightly lower S2S Score (3.31), consistent with the view that the SFT checkpoint already captures most of the layout knowledge and DPO mainly fine-tunes timing.

\paragraph{Effect of Full-Duplex Training on Response Quality}
A natural concern is that learning timing decisions might erode the underlying response quality. We compare \name (SFT) with a turn-based SFT baseline trained on the same data in the original GLM-4-Voice format. Table~\ref{tab:fullduplex_turnbased} shows that the duplex model is on par with or stronger than the turn-based one: it loses 1.0 point on Llama Questions but gains 2.1 points on Web Questions and 0.07 on Alpaca-Eval. This indicates that multi-channel interleaved training introduces full-duplex behavior without sacrificing response quality: the gains in turn-taking and interruption come from a layout that exposes timing as an in-vocabulary prediction problem, not from a degraded language model.

\section{Related Work}
\label{sec:related}

\paragraph{Speech Language Models}
SpeechLMs are generally divided into two categories: native SpeechLMs that directly input and output speech tokens through a decoder-only Transformer (SpeechGPT~\citep{zhang2023speechgpt}, GLM-4-Voice~\citep{glm4voice}, IntrinsicVoice~\citep{zhang2024intrinsicvoice}, Spirit-LM~\citep{nguyen2024spirit}, Step-Audio~\citep{huang2025stepaudio,wu2025stepaudio2}), and modular SpeechLMs that add speech encoders and decoders around the LLM (LLaMA-Omni~\citep{fang2024llamaomni}, LLaMA-Omni 2~\citep{fang2025llamaomni2}, Mini-Omni~\citep{xie2024miniomni}, SALMONN~\citep{tang2024salmonn}, Freeze-Omni~\citep{wang2024freezeomni}, MinMo~\citep{chen2025minmo}, Stream-Omni~\citep{zhang2025streamomni}, VITA-1.5~\citep{fu2025vita15}, VITA-Audio~\citep{long2025vitaaudio}). Native models inherit the LLM training stack with minimal architectural changes, but they enlarge the per-step softmax with the union of text and speech tokens and require continued pretraining on large amounts of speech to keep the model's text capability from collapsing. Modular models keep the LLM vocabulary clean and reuse strong off-the-shelf speech encoders and decoders, at the cost of a more elaborate training pipeline that must align the inserted modules with the frozen or partially-trained LLM. Both families assume that a complete user utterance is available before the model speaks, and segment the user audio with an external VAD; \name removes the VAD entirely and lets the model itself decide when to speak.

\paragraph{Full-Duplex Speech Language Models}
Full-duplex SpeechLMs lift the turn-based assumption. dGSLM~\citep{nguyen2022dgslm} pioneered dual-channel modeling on naturalistic conversational speech, demonstrating that a single autoregressive model can predict both speakers without an external turn-taking signal, but at the cost of relying on tens of thousands of hours of two-channel dialogue and offering limited semantic coverage. Moshi~\citep{defossez2024moshi} folds the user and assistant audio into two parallel residual vector quantization (RVQ) streams stacked over a text Inner Monologue and uses a depth-Transformer to emit one frame per step for low theoretical latency; the parallel-RVQ design requires per-codebook conditioning and full-duplex pretraining at the scale of millions of hours of speech. SyncLLM~\citep{veluri2024syncllm} embeds an explicit wall-clock signal so that user and assistant tokens advance in lock-step, but the time tokens enlarge the vocabulary and shift the burden of timing to the LLM. OmniFlatten~\citep{zhang2024omniflatten} flattens user-speech, assistant-speech, and assistant-text tokens into a single GPT stream, which simplifies the training stack but interleaves channels at the per-token granularity and fragments the contiguous text monologue. SALMONN-omni~\citep{yu2024salmonnomni} runs on continuous embeddings with a thinking mechanism, sidestepping the discretization trade-offs but requiring a separate codec for the audio output and an extra branch for the thinking trace. LSLM~\citep{ma2024lslm}, Mini-Omni2~\citep{xie2024miniomni2}, and Freeze-Omni~\citep{wang2024freezeomni} reach partial duplexity through input-side barge-in or command-based interruption: the model can be cut off but cannot decide for itself when to start or stop talking. \citet{zhang2024duplex} reach duplexity at the text level via time-division multiplexing. The most relevant concurrent work is FLM-Audio~\citep{yao2025flmaudio}, which similarly preserves natural text monologues but merges all channels at every step. \name interleaves three channels at a coarse block granularity that preserves contiguous text monologues, and unifies all dialogue-state decisions as next-token prediction over the standard LLM vocabulary.

\paragraph{Text Channels and Inner Monologues}
Many full-duplex SpeechLMs introduce an intermediate text channel as scaffolding for speech generation. Moshi~\citep{defossez2024moshi} interleaves a per-frame Inner Monologue track that emits time-aligned text before each frame of audio and reports that this track is critical for keeping the spoken response semantically coherent. SALMONN-omni~\citep{yu2024salmonnomni} pursues a similar idea on continuous embeddings with a separate thinking branch. OmniFlatten~\citep{zhang2024omniflatten} and FLM-Audio~\citep{yao2025flmaudio} likewise weave text alongside speech tokens, reusing the LLM's text generation pathway to plan content. Our text channel inherits this design at the granularity of one block rather than one frame: it never reaches the user but conditions the assistant-speech tokens within the same block, and it is the channel where every dialogue-state decision is made. Compared with per-step interleaving, the coarser scheme keeps each utterance's text contiguous over several consecutive blocks, which we conjecture aligns better with the text distribution that the underlying LLM was pretrained on.

\section{Conclusion}
\label{sec:conclusion}

We introduce \name, a native full-duplex SpeechLM whose multi-channel interleaved sequence lets a single autoregressive LLM decide when to listen, speak, and stop. Four dialogue-state tokens added to the standard vocabulary turn turn-taking and interruption into ordinary next-token prediction, with no auxiliary classifier or scheduler on top of GLM-4-Voice. With only 400K full-duplex samples and a lightweight DPO stage, \name reaches 92\% turn-taking and 100\% interruption success.

\section*{Limitations}

The training and evaluation audio is fully synthesized: it is single-speaker, near-field, and noise-free. Real-world deployment must handle background noise, reverberation, and competing speakers, all of which can shift the boundaries detected in the user channel and trigger spurious turn-taking events. We expect data augmentation (additive noise, room impulse response, distractor speakers) to mitigate this, and we leave a controlled study across in-car, outdoor, and meeting-room conditions to future work. Our analysis also focuses on turn-taking and interruption; backchannels, multi-party conversation, and emotion-aware turn-taking are not explored. The chosen block size $N{=}10$ caps the minimum response latency at 0.8\,s; reducing $N$ would lower latency but shrink the per-block text budget, and we leave a systematic sweep over $N$ to future work. Finally, like Moshi and OmniFlatten, our model is bounded by the quality and the bias of the underlying SpeechLM (GLM-4-Voice); we share its limitations on rare languages, code-switching, and out-of-distribution acoustic conditions.

\section*{Ethical Considerations}

\name{} synthesizes natural-sounding speech in real time, which lowers the barrier for voice-based impersonation, social-engineering attacks, and audio disinformation. Continuous-listening interfaces also raise privacy concerns: always-on user-channel input may inadvertently capture private speech, including utterances from bystanders who have not consented to recording. As \name{} is built on top of GLM-4-Voice, it inherits the linguistic, demographic, and acoustic biases of that backbone, and its turn-taking and interruption decisions may behave unevenly across speakers, accents, and languages. We release the model strictly for research on full-duplex dialogue modeling; production deployments should add speaker verification, on-device wake-word gating, watermarking of synthesized speech, and explicit user consent for continuous capture.

\bibliography{custom}

\end{document}